%% file: main.tex
\documentclass{article}

\usepackage[final,nonatbib]{neurips2020_preregistration}

\usepackage[utf8]{inputenc} 
\usepackage[T1]{fontenc}    
\usepackage{hyperref}       
\usepackage{url}            
\usepackage{booktabs}       
\usepackage{amsfonts}       
\usepackage{nicefrac}       
\usepackage{microtype}      
\usepackage{amsmath}
\usepackage{color}
\usepackage{enumitem}

\title{
An Empirical Study of Explainable AI Techniques\\ on Deep Learning Models For Time Series Tasks
}

\author{%
  Udo Schlegel\\
  University of Konstanz\\
  \texttt{u.schlegel@uni-konstanz.de} \\
   \And
   Daniela Oelke \\
   Offenburg University of Applied Sciences \\
   \texttt{daniela.oelke@hs-offenburg.de} \\
   \AND
   Daniel A. Keim \\
   University of Konstanz \\
   \texttt{keim@uni.kn} \\
   \And
   Mennatallah El-Assady \\
   University of Konstanz \\
   \texttt{menna.el-assady@uni.kn} \\
}

\begin{document}

\maketitle

\input{content/0-abstract.tex}
\input{content/1-introduction.tex}
\input{content/2-related-work}
\input{content/3-methodology}
\input{content/4-conclusion}

\begin{ack}
This work was partly funded by the European Union’s Horizon 2020 research and innovation programme under grant agreements No 826494.
\end{ack}

\bibliographystyle{abbrv}
\bibliography{main}

\end{document}

%% file: content/0-abstract.tex
\vspace{-0.5cm}
\begin{abstract}
Decision explanations of machine learning black-box models are often generated by applying Explainable AI (XAI) techniques. 
However, many proposed XAI methods produce unverified outputs.
Evaluation and verification are usually achieved with a visual interpretation by humans on individual images or text.
In this preregistration, we propose an empirical study and benchmark framework to apply attribution methods for neural networks developed for images and text data on time series.
We present a methodology to automatically evaluate and rank attribution techniques on time series using perturbation methods to identify reliable approaches.
\end{abstract}

%% file: content/1-introduction.tex
\vspace{-0.2cm}
\section{Introduction}
Explainable AI (XAI) establishes a research field to bridge the gap between state-of-the-art deep learning and production-ready models in the industrial sector to explain and understand outputs.
In research, deep learning achieves state-of-the-art performance in autonomous driving~\cite{huval_empirical_2015}, speech assistance~\cite{dahl_context-dependent_2011}, and natural language processing~\cite{vaswani_attention_2017} improving previous results by a margin.
However, deep neural network models' black-box nature is often not suitable for a production-ready model~\cite{gunning_d._explainable_2016}.
For instance, the EU General Data Protection Regulation~\cite{european_union_european_2018} forces companies to justify their employed algorithms' decisions.
Especially, critical characteristics (fairness, privacy, reliability, trust~\cite{doshi-velez_towards_2017}) for machine learning systems led to the dismissal of state-of-the-art deep learning models and the employment of interpretable models, e.g., in critical environments like health~\cite{rudin_please_2018}.
However, agencies like DARPA promoted research grants around explainable AI projects~\cite{gunning_d._explainable_2016} to support the research into understanding deep learning models to facilitate their deployment.
XAI techniques such as local interpretable model-agnostic explanations (LIME)~\cite{ribeiro_why_2016} are developed to enable the interpretation of the model's decisions~\cite{guidotti_survey_2018}.
These XAI techniques promise to bridge the gap between black-box decision and comprehensible explanations of such models.

Most XAI techniques aiding in debugging and understanding a model's decisions are generally developed for domains such as images or text~\cite{guidotti_survey_2018}.
However, other types of data emerge as fast as images or text, e.g., due to an increasing amount of installed sensors.
Such sensors produce massive amounts of time series signals and support, for instance, early detection of failures through failure prediction models~\cite{mobley_introduction_2002}.
An increasing amount of black-box models are deployed to tackle tasks with time series data.
However, only a few works include temporal data in their XAI technique analysis, e.g., LIME~\cite{ribeiro_why_2016}, to create comprehensible explanations for such complex models.
Such a development urges to evaluate already established XAI techniques applied on time series to identify applicable methods~\cite{schlegel_towards_2019}.
As time series analysis is often domain and tasks specific, a rigorous study of XAI techniques applied on such broad on time series can identify strengths and weaknesses to analyze baseline functionality for a time series XAI method~\cite{schlegel_towards_2019}.

In many cases, evaluating XAI techniques on a large scale is a tedious task, as a human evaluation is typical for most domains, which is relatively slow and sample-based~\cite{mohseni_survey_2018}.
However, a quantitative evaluation is necessary to identify reliable methods and verify their explanations on the number of data sensors produce.
Thus, automatic quantitative analysis is needed to support human qualitative evaluations to present only critical samples for further investigation of the model and the XAI technique.
Also, raw time series and sensor signals are often difficult for domain experts to analyze without converting them.
Thus verification of explanations based on raw time series is not trivial even with human evaluations.
In many cases, Fourier transformations help analysts to understand data characteristics. 
Still, with complex models working on the raw data, such a transformation often comprises loss of information, biases, or even faulty conclusions.
As a result, automated evaluation and verification of XAI techniques on complex raw time series models are required to support humans in understanding, debugging, and improving their models.

In this preregistration, we propose an empirical study and benchmark framework to test and evaluate XAI techniques, focusing on attribution methods applied to time series data and models, thereby extending our previous work~\cite{schlegel_towards_2019}.
We present a methodology to automatically evaluate and rank XAI techniques on time series tasks using perturbation methods.
Our empirical analysis tackles the following questions about the verification of XAI techniques on a selected set of network architectures.

\textbf{Transferring attribution methods to time series:}
\vspace{-0.3\baselineskip}
\begin{itemize}[noitemsep,topsep=0pt]
    \item For each of the considered XAI attribution methods, can we transfer them to time series?
    \item What are the tweaks and changes needed to apply each method? 
    \item Are some methods better suited for particular models and tasks?
\end{itemize}
\textbf{Evaluating the applicability of transferred methods to time series tasks:}
\vspace{-0.3\baselineskip}
\begin{itemize}[noitemsep,topsep=0pt]
    \item Which metrics and measures of validation are required for a systematic evaluation of each method on the given tasks?
    \item How strong does an XAI attribution reflect the model's predictions? 
    \item Can we rank and identify strengths as well as weaknesses of attributions on time series for given tasks?
\end{itemize}
\textbf{Benchmarking attribution methods on time series tasks:}
\vspace{-0.3\baselineskip}
\begin{itemize}[noitemsep,topsep=0pt]
    \item Given a concrete model architecture, dataset, and task; how sensitive are XAI methods?
    \item What is the overall strongest performing XAI method for each task?
    \item Can we reproduce the results of Schlegel et al.~\cite{schlegel_towards_2019} showing SHAP as best performing?
\end{itemize}
We want to investigate how much the XAI technique outputs differ from each other and how much their proposed domain differs from time series tasks through these questions to investigate a detailed analysis of strengths and weaknesses.
Such an analysis leads to insights into the model's behaviors towards the time dependencies these complex models learn during their training.

%% file: content/2-related-work.tex
\section{Related Work}
An increasing amount of XAI techniques are developed to support, reason, and explain the rising amount of AI model decisions~\cite{spinner_explainer_2019}.
Especially, methods like LIME~\cite{ribeiro_why_2016} and Integrated Gradients~\cite{shrikumar_learning_2017} slowly begin to move into industrial contexts to facilitate the usage of production-ready deep learning models. 
Nevertheless, most XAI techniques are only evaluated on their data and task at hand to show their applicability~\cite{sundararajan_axiomatic_2017, shrikumar_learning_2017}. 
In some cases, new XAI techniques are compared to some others. 
However, these comparisons are relatively sparse and only against a few similar techniques~\cite{kapishnikov_xrai_2019}. 
There is often a lack in comparison to more than just similar methods and techniques.
Further, general evaluation is often only done using, e.g., the pointing game for object detection~\cite{zhang_top-down_2018} which checks for every object if the maximum attribution of the XAI technique is contained in the bounding box.

Nevertheless, Hooker et al.~\cite{hooker_benchmark_2019} propose the benchmark framework ROAR (RemOve And Retrain) on images to investigate XAI techniques and their feature attributions.
ROAR first uses the attributions to perturb relevant pixels to uninformative values to create a new data set.
Next, it tests the model's accuracy on the created data and retrains the model on it.
After the retraining, ROAR again creates attributions and perturbs the relevant pixels to get a new data set to test the accuracy.
Hooker et al.~\cite{hooker_benchmark_2019} find that most of their applied techniques produce unconvincing attributions on the retrained model which do not change the test accuracy.
The benchmark has a few weaknesses, for instance, complex models, which need a lot of time for their training, are hard to analyze as the whole benchmark run needs to train the model twice which could lead to days or weeks for results.
Also, ROARs results show that after retraining the model on the changed data, the accuracy does not get worse by perturbing the data by applying the XAI technique again.
Such a result show that these techniques not always show the correct explanations, however, these results could also show that the XAI technique is not suitable for the model architecture.

Schlegel et al.~\cite{schlegel_towards_2019} propose two verification methods to apply perturbation methods onto time series data and present preliminary results applying XAI techniques on time series.
They propose time point perturbations, which change relevant time points to zero or the original value's inverse.
To test the time dependencies learned by a classifier, they present time interval perturbations in which they reorder an interval around relevant time points or set the whole interval to the its mean.
Through these verification methods, Schlegel et al.~\cite{schlegel_towards_2019} demonstrate in a preliminary evaluation the application of  Saliency Maps~\cite{simonyan_deep_2013}, LRP~\cite{bach_pixel-wise_2015}, DeepLIFT~\cite{shrikumar_learning_2017}, LIME~\cite{ribeiro_why_2016}, and SHAP~\cite{lundberg_unified_2017} on time series.
However, they neglect other XAI techniques to investigate the verification methods they propose and show the results only for one gradient-based approach.
Further, an analysis between score-based and gradient-based techniques could lead to strengths and weaknesses for specific models.

Overall, there is a lack of XAI technique evaluations as the research is still only applied in some areas like computer vision~\cite{hooker_benchmark_2019, kapishnikov_xrai_2019}.
Thus, in this work, we tackle the time series domain to apply and analyze other fields' XAI techniques to evaluate suitability and explanations.

%% file: content/3-methodology.tex
\section{Methodology}
To assess the quality of an explanation of an XAI technique, we measure the difference of a quality metric applied on the models output for test data and on the output of an adaption of the test data based on the explanation.

\subsection{Perturbation on Time Series}
One of the most used techniques to test impacts on the classification of models are perturbations of the data~\cite{schlegel_towards_2019,zeiler_visualizing_2014}.
A common technique in computer vision to either evaluate a model on the importance of pixels in an image is to set these to black~\cite{zeiler_visualizing_2014} or test an XAI technique by adapting the relevant pixels to an uninformative value (e.g., mean)~\cite{hooker_benchmark_2019}.
Schlegel et al.~\cite{schlegel_towards_2019} proposed two verification methods to use similar perturbation techniques on time series as altering a time point to zero often leads to biased data, which can help to classify.
Such perturbations at single time points do not test if a classifier learns just single time points or if time dependencies are learned to predict unseen data.
Thus, Schlegel et al.~\cite{schlegel_towards_2019} propose a verification method on time intervals around relevant time points in which they reorder the interval or set the interval to the mean of it.

In our experiment, we adapt the proposed verification methods of Schlegel et al.~\cite{schlegel_towards_2019} and add some tweaks to extend these to seven methods.
For our time point perturbation, we take the relevant time points and set them to zero, to their inverse, and to the mean of the time series sample.
The used perturbation takes a time series $t = (t_{0}, t_{1}, t_{2}, ..., t_{n})$ and changes relevant time points to, e.g., zero resulting in the changed time series $t^c = (t_{0}, 0, t_{2}, ..., 0, t_{n})$ for relevant time points $i = 1, n-1$.
Our time interval perturbation sets an interval around a relevant time point to zero, to the mean of the interval, to the inverse of every time point interval, and to a reordered version like Schlegel et al.~\cite{schlegel_towards_2019}.
These interval perturbations differ from the point perturbation by changing a whole intervals in a time series $t = (t_{0}, t_{1}, t_{2}, ..., t_{n})$ to, for instance, the mean of the interval resulting in $t = (t_{0}, \mu_{2}, \mu_{2}, \mu_{2}, ..., t_{n})$ with a relevant time point at $i = 2$ and an interval range of $1$.
Through these types of perturbations, we achieve a divers range of testing setups with various possible hypothesis.
For instance, we can analyze models and XAI techniques on key features, e.g., models focusing on time intervals rather than points or techniques showing only time points as relevant.

\subsection{Evaluation Methodology}
After defining our verification methods, our evaluation methodology consists in the first step of the selection process for a task, a dataset, and a model (e.g., task: forecasting, data: weather, and model: RNN with LSTM~\cite{hochreiter_long_1997} layers).
Our evaluation pipeline is composed of three stages (model training, model explanation, and explanation evaluation) to calculate a score for every XAI technique we consider for the model architecture.

\textbf{Model training --}
In the first step, we train our selected architecture on the selected data with the assigned task with a chosen cost function, e.g., RMSE.
The model is initialized with a static seed and is trained for a previously chosen amount of epochs, neglecting early stopping to have a reproducible outcome.
If the model's training time does not exceed a specific limit, the process is changed to a random seed and ten trained models.
In this case, the score is calculated by the average of the ten models.
Finally, we use the quality metric to evaluate the XAI technique on a test set to get a baseline.
Such a baseline can consist, for instance, of the accuracy or the RMSE depending on the task.

\textbf{Model explanation --}
After training the model, we apply our XAI techniques on every sample of the test data and calculate the attributions.
We normalize the attributions on single samples to identify the most relevant time points.
Next, we select relevant time points for verification of our test data in three ways.
First, we sort the resulting attributions of every sample and take the top k attributions.
The sorting helps us to get the positions of the most relevant k time points.
The parameter k varies with the input length and is relatively chosen to be five percent of the length with $len * 0.05$.
Second, we calculate a threshold with $max - (max - mean) * 0.1$ of the attributions and take all time points with an attribution above as relevant.
Such a threshold leads to the 95 percentile if the attributions reflect a normal distribution; else, it leads to a more robust selection as it is more dynamic than the top k.
Third, we take a fixed threshold (e.g., $0.8$) and select all time points with a higher attribution value than the threshold as relevant time points. 
After the selection of points, we perturb the selected time points at the resulting attribution positions for all three processes according to our verification methods.
These three perturbations lead to $3 * v$ new test sets $t^{c}$ with $v$ being the number of verification methods applied and $c$ an identifier for the change.
As a last, we create again $3 * v$ new test sets $t_r^{c}$ by selecting time points randomly as relevant based on the number of relevant points by the previous selections as well as ten more and ten less percent.
Such three random baselines enable better comparison options.

\textbf{Explanation evaluation --}
In our last step, we take our newly $12 * v$ created test sets $t^{c}$ as well as $t_r^{c}$ and our model to predict the quality metric to compare the results to the baseline.
The assumption $qm(t)\geq qm(t_r^{c}) > qm(t^{c})$, with the quality measure $qm$, the test set $t$, the randomly changed points $t_r^{c}$, and the XAI relevant time points $t^{c}$, holds, if the XAI technique captures relevant information the model learns to distinguish samples.

\section{Experimental Protocol}
Our current experimental setting involves the usage of 15 XAI techniques, six DNN models, seven verification techniques, and three time series tasks with various data sets.
As such a setting, our design space of techniques and algorithms consists of at least 15 * 6 * 7 * 3 possible variations.
However, as the setup will also be provided as an extendable benchmark framework, more XAI techniques, models, data sets, and verification methods can be added later on to test further parameters.

\noindent\textbf{XAI Techniques --}
The XAI techniques will involve gradient, score, and surrogate based attribution methods, namely LRP~\cite{bach_pixel-wise_2015}, LIME~\cite{ribeiro_why_2016}, SHAP~\cite{lundberg_unified_2017}, IG~\cite{sundararajan_axiomatic_2017}, Saliency~\cite{simonyan_deep_2013}, Occlusion~\cite{zeiler_visualizing_2014}, DeepLIFT~\cite{shrikumar_learning_2017}, Input*Grad~\cite{shrikumar_not_2016}, PatternAttribution~\cite{zeiler_visualizing_2014}, Shapely Sampling~\cite{castro_polynomial_2009}, GradCAM~\cite{selvaraju_grad-cam_2017}, Guided Backprop~\cite{springenberg_striving_2014}, NoiseTunnel~\cite{adebayo_sanity_2018}, DeepTaylor~\cite{montavon_explaining_2017}, and SmoothGrad~\cite{smilkov_smoothgrad:_2017}.
We will use default parameters for all XAI techniques or will use parameters, as mentioned in the original papers.
For our benchmark framework, we will also provide functions to evaluate and test different parameters, e.g., changing the kernel size for LIME or the window size for Occlusion.

\noindent\textbf{DNN Models --}
Our models will consist of various deep neural networks with recurrent neural networks (LSTM~\cite{hochreiter_long_1997}, GRU~\cite{cho_learning_2014}), convolutional neural networks (Conv1D, Conv2D, Conv3D), transformers for time series~\cite{li_enhancing_2019, vaswani_attention_2017}, and a deep neural network with residual connections~\cite{he_deep_2016}.
We will initialize our models with uniform distribution of weights and biases near zero.
The models will be trained using Adam~\cite{kingma_adam:_2014} as a optimizer.
Our benchmark framework will also enable to change the models to predefined architectures with pre-trained parameters.

\noindent\textbf{Verification Techniques --}
As a base for the verification, we include the methods presented in the previous section, namely point perturbations (zero, inverse, mean) and interval perturbations (swap, mean, zero, inverse) with accuracy (1) and root mean squared error (2) as deployed quality measures.

\noindent\begin{minipage}{.5\linewidth}
\begin{equation}
  ACC = \frac{\text{correct predictions}}{\text{amount of samples}}
\end{equation}
\end{minipage}%
\begin{minipage}{.5\linewidth}
\begin{equation}
  RMSE = \sqrt{\frac{1}{n}\Sigma_{i=1}^{n}{(y_i - \hat{y}_i)^2}}
\end{equation}
\end{minipage}

Accuracy has a problem with unbalanced data sets as a model could only learn to predict the majority class and the accuracy would still be good.
However, in our case, even if it learns to predict the majority class every time, we can argue that a working attribution should find relevant time points.
Even if our classifier predicts the same for everything, then every XAI technique (even the random change) should have zero impact on the classification, which indicates a flawed model.

\noindent\textbf{Time Series Tasks --}
We will focus on three major time series tasks for our experiment setting, namely classification, forecasting, and regression.
As there is often ambiguity in these terms, we define them as follows:
A classification model labels segmented non-overlapping time series $t$ with a class $c$.
A forecasting model slides over a time series $t$ and predicts for every step a separate value $y$ not included in the time series.
A regression model predicts the next time point $t_{i+1}$ or interval based on the previous time series $t_{0,...,i}$.
For the classification tasks, we will use the UCR Benchmark repository~\cite{dau_ucr_2018} with 128 different data sets.
Our forecasting data set will consist of weather forecasting~\footnote{Deutscher Wetterdienst: https://opendata.dwd.de/}, air quality~\cite{zhang_cautionary_2017}, and store demand forecasting~\cite{dua_uci_2017}.
And lastly, our regression task will incorporate finance data~\footnote{Yahho Finance: https://finance.yahoo.com/quotes/OCR,dataset/view/v1/} and household power consumption~\cite{dua_uci_2017}.

\noindent\textbf{Implementation --}
For our experiment, we will use PyTorch~\footnote{PyTorch Deep Learning Framework: https://pytorch.org/} and Captum~\footnote{Captum Model Interpretability for PyTorch: https://captum.ai/} as primary libraries.
PyTorch will be used to build and train the selected models on the selected data sets.
Captum will create the attributions we selected to analyze based on the trained PyTorch model.
In some cases, like LIME and SHAP, we will use the implementations provided by the authors.
We extend the Pytorch LRP implementation for LSTMs with the implementation provided by Arras et al.~\cite{arras_evaluating_2019}.
For reproducibility, we fix all possible random seeds to 13.

\noindent\textbf{Experiment Run --}
Our experiment run will consist of independent docker containers that apply our methodology for a selected time series task, data set, and model.
Such a configuration enables to fix a time series dataset and a model to investigate the various attribution methods.
We run our experiments strictly using the previously presented methodology using our three-stage evaluation.
After a successful run, the container will store the trained model and the results in an external database.
We can achieve a highly efficient experiment setup through the independent docker containers as hardware constraints of our servers only limit us.

\noindent\textbf{Result Analysis --}
After our experiment run, we will analyze the results to investigate the XAI techniques' strengths and weaknesses towards architectures and tasks to come up with a best-practices guideline.
Based on these findings, we will either prove or deny general views about XAI techniques applied on other data than images or text answering the questions we brought up in previously.
We will also focus on the previously presented exploratory questions and investigate exploratory tasks to identify surprising results. 
A closer look into these questions enables us to enhance our preregistration with a section of unforeseen outcomes as related work misses hypotheses about the XAI techniques.
However, we will reproduce our preliminary results~\cite{schlegel_towards_2019} and investigate our previous hypothesis of SHAP being the overall best XAI technique.
Our extended set of random baselines will facilitate either prove or reject a significance towards this hypothesis.
With this preregistration and the exploratory results and findings, we will establish a useful benchmark for attribution techniques on time series to foster the research into novel methods.

%% file: content/4-conclusion.tex
\section{Extension as a benchmark framework}
As new XAI techniques are getting developed rapidly, we will extend our methodology and experimental setup into a benchmark framework to support novel approaches' analysis and verification.
By enabling users to add their own models, data sets, verification methods, and XAI techniques, the benchmark framework supports to analyze and compare novel methods and models against prominent and state-of-the-art techniques.
We will further extend the framework with the concept of ROAR (RemOve And Retrain) by Hooker et al.~\cite{hooker_benchmark_2019} to enhance the functionality of the benchmark and to be able to also test XAI techniques before production ready models. 
Through such an extension, our framework will be able to test combinations of XAI techniques to dig deep into the interplay of these.
Good working interplays lead to robust explanations of ensembles of XAI techniques.